\documentclass[10pt]{article}

\usepackage{amsmath}
\usepackage{amssymb}
\usepackage{graphicx}
\usepackage{cite}
\usepackage[usenames,dvipsnames]{color}
\usepackage[normalem]{ulem}
\usepackage{epsfig}
\usepackage{subfig}
\usepackage{float}
\usepackage{dsfont}
\usepackage{xcolor}
\usepackage[normalem]{ulem}
\usepackage{color}
\usepackage{algorithm}
\usepackage{algorithmic}
\usepackage{booktabs}
\usepackage{hyperref}
\usepackage{placeins}

\usepackage[labelfont=bf,labelsep=period,justification=raggedright]{caption}

\makeatletter
\renewcommand{\@biblabel}[1]{\quad#1.}
\makeatother

\date{}

\definecolor{col}{rgb}{0.9,0.0,0.0}

\newcommand{\cl}[1]{#1}

\begin{document}
\begin{flushleft}
{\Large
\textbf{``What is Relevant in a Text Document?'':\\ An Interpretable Machine Learning Approach}
}
\\
Leila Arras$^{1}$, 
Franziska Horn$^{2}$, 
Gr{\'e}goire Montavon$^{2}$,
Klaus-Robert M{\"u}ller$^{2,3,*}$,
Wojciech Samek$^{1,*}$
\\
\bf{1} Machine Learning Group, Fraunhofer Heinrich Hertz Institute, Berlin, Germany
\\
\bf{2} Machine Learning Group, Technische Universit\"at Berlin, Berlin, Germany
\\
\bf{3} Department of Brain and Cognitive Engineering, Korea University, Seoul, Korea
\\
$\ast$ Email: klaus-robert.mueller@tu-berlin.de, wojciech.samek@hhi.fraunhofer.de
\end{flushleft}

\begin{abstract}
Text documents can be described by a number of abstract concepts such as semantic category, writing style, or sentiment. Machine learning (ML) models have been trained to automatically map documents to these abstract concepts, allowing to annotate very large text collections, more than could be processed by a human in a lifetime. 
Besides predicting the text's category very accurately, it is also highly desirable to understand {\it how} and {\it why} the categorization process takes place. 
In this paper, we demonstrate that such understanding can be achieved by tracing the classification decision back to individual words using layer-wise relevance propagation (LRP), a recently developed technique for explaining predictions of complex non-linear classifiers.
We train two word-based ML models, a convolutional neural network (CNN) and a bag-of-words SVM classifier, on a topic categorization task and adapt the LRP method to decompose the predictions of these models onto words. Resulting scores indicate how much individual words contribute to the overall classification decision. This enables one to distill relevant information from text documents without an explicit semantic information extraction step.
We further use the word-wise relevance scores for generating novel vector-based document representations which capture semantic information. Based on these document vectors, we introduce a measure of {\it model explanatory power} and show that, although the SVM and CNN models perform similarly in terms of classification accuracy, the latter exhibits a higher level of explainability which makes it more comprehensible for humans and potentially more useful for other applications.

\end{abstract}

\section{Introduction}

A number of real-world problems related to text data have been studied under the framework of natural language processing (NLP). Example of such problems include topic categorization, sentiment analysis, machine translation, structured information extraction, or automatic summarization. Due to the overwhelming amount of text data available on the Internet from various sources such as user-generated content or digitized books, methods to automatically and intelligently process large collections of text documents are in high demand.
For several text applications, machine learning (ML) models based on global word statistics like TFIDF \cite{IDF_Jones, TFIDF_Salton} or linear classifiers are known to perform remarkably well, e.g.\ for unsupervised keyword extraction \cite{Hasan} or document classification \cite{Aggarwal}. However more recently, neural network models based on vector space representations of words (like \cite{Mikolov2}) have shown to be of great benefit to a large number of tasks. The trend was initiated by the seminal work of \cite{Bengio} and \cite{Collobert}, who introduced word-based neural networks to perform various NLP tasks such as language modeling, chunking, named entity recognition, and semantic role labeling. A number of recent works (e.g.\  \cite{Collobert, Socher-etal:2013}) also refined the basic neural network architecture by incorporating useful structures such as convolution, pooling, and parse tree hierarchies, leading to further improvements in model predictions. Overall, these ML models have permitted to assign automatically and accurately concepts to entire documents or to sub-document levels like phrases; the assigned information can then be mined on a large scale.

In parallel, a set of techniques were developed in the context of image categorization to explain the predictions of convolutional neural networks (a state-of-the-art ML model in this field) or related models. These techniques were able to associate to each prediction of the model a meaningful pattern in the space of input features \cite{DBLP:conf/eccv/ZeilerF14, Simonyan, SchuettArXiv16} or to perform a decomposition onto the input pixels of the model output\cite{DBLP:conf/cidm/LandeckerTBMKB13, Bach, MonArXiv15}. In this paper, we will make use of the layer-wise relevance propagation (LRP) technique \cite{Bach}, that was already substantially tested on various datasets and ML models \cite{SamArXiv15, ArbGCPR16, StuJNM16, BacCVPR16}.

In the present work, we propose a method to identify which words in a text document are important to explain the category associated to it. The approach consists of using a ML classifier to predict the categories as accurately as possible, and in a second step, decompose the ML prediction onto the input domain, thus assigning to each word in the document a relevance score. The ML model of study will be a word-embedding based convolutional neural network that we train on a text classification task, namely topic categorization of newsgroup documents. As a second ML model we consider a classical bag-of-words support vector machine (BoW/SVM) classifier.
 
\vspace*{0.25cm}
\noindent We contribute the following:\\[+3px]
(i) The LRP technique \cite{Bach} is brought to the NLP domain and its suitability for identifying relevant words in text documents is demonstrated.\\[+3px]
(ii) LRP relevances are validated, at the document level, by building document heatmap visualizations, and at the dataset level, by compiling representative words for a text category. 
It is also shown quantitatively that LRP better identifies relevant words than sensitivity analysis.\\[+3px]
(iii) A novel way of generating vector-based document representations is introduced and it is verified that these document vectors present semantic regularities within their original feature space akin word vector representations.\\[+3px]
(iv) A measure for model explanatory power is proposed and it is shown that two ML models, a neural network and a BoW/SVM classifier, although presenting similar classification performance may largely differ in terms of explainability.

\vspace*{0.25cm}
The work is organized as follows. In section \ref{sec2} we describe the related work for explaining classifier decisions with respect to input space variables. In section \ref{sec3} we introduce our neural network ML model for document classification, as well as the LRP decomposition procedure associated to its predictions. We describe how LRP relevance scores can be used to identify important words in documents and introduce a novel way of condensing the semantical information of a text document into a single document vector. Likewise in section \ref{sec3} we introduce a baseline ML model for document classification, as well as a gradient-based alternative for assigning relevance scores to words.
In section \ref{sec4} we define objective criteria for evaluating word relevance scores, as well as for assessing model explanatory power.
In section \ref{sec5} we introduce the dataset and experimental setup, and present the results. Finally, section \ref{sec6} concludes our work.

\section{Related Work}\label{sec2}
Explanation of individual classification decisions in terms of input variables has been studied for a variety of machine learning classifiers such as additive classifiers \cite{Poulin:2006:VEE:1597122.1597143}, kernel-based classifiers \cite{BaehrensSHKHM10} or hierarchical networks \cite{DBLP:conf/cidm/LandeckerTBMKB13}. Model-agnostic methods for explanations relying on random sampling have also been proposed \cite{Strumbelj10, DBLP:conf/kdd/Ribeiro0G16, TurnerArXiv16}. Despite their generality, the latter however incur an additional computational cost due to the need to process the whole sample to provide a single explanation.
Other methods are more specific to deep convolutional neural networks used in computer vision: the authors of \cite{DBLP:conf/eccv/ZeilerF14} proposed a network propagation technique based on deconvolutions to reconstruct input image patterns that are linked to a particular feature map activation or prediction. The work of \cite{Simonyan} aimed at revealing salient structures within images related to a specific class by computing the corresponding prediction score derivative with respect to the input image. The latter method reveals the sensitivity of the classifier decision to some \textit{local variation} of the input image, and is related to sensitivity analysis \cite{Dimopoulos95, Gevrey03}.
In contrast, the LRP method of \cite{Bach} corresponds to a {\it full decomposition} of the classifier output for the current input image. It is based on a layer-wise conservation principle and reveals parts of the input space that either support or speak against a specific classification decision. 
Note that the LRP framework can be applied to various models such as kernel support vector machines and deep neural networks \cite{Bach, BacCVPR16}. 
We refer the reader to \cite{SamArXiv15} for a comparison of the three explanation methods, and to \cite{MonArXiv15} for a view of particular instances of LRP as a ``deep Taylor decomposition'' of the decision function. 

In the context of neural networks for text classification \cite{Denil2} proposed to extract salient sentences from text documents using loss gradient magnitudes. In order to validate the pertinence of the sentences extracted via the neural network classifier, the latter work proposed to subsequently use these sentences as an input to an external classifier and compare the resulting classification performance to random and heuristic sentence selection.
The work by \cite{Li} also employs gradient magnitudes to identify salient words within sentences, analogously to the method proposed in computer vision by \cite{Simonyan}. However their analysis is based on qualitative interpretation of saliency heatmaps for exemplary sentences. In addition to the heatmap visualizations, we provide a classifier-intrinsic quantitative validation of the word-level relevances. We furthermore extend previous work from \cite{arras-acl16} by adding a BoW/SVM baseline to the experiments and proposing a new criterion for assessing model explanatory power.

\section{Interpretable Text Classification}\label{sec3}

\begin{figure}
\includegraphics[width=0.95\textwidth]{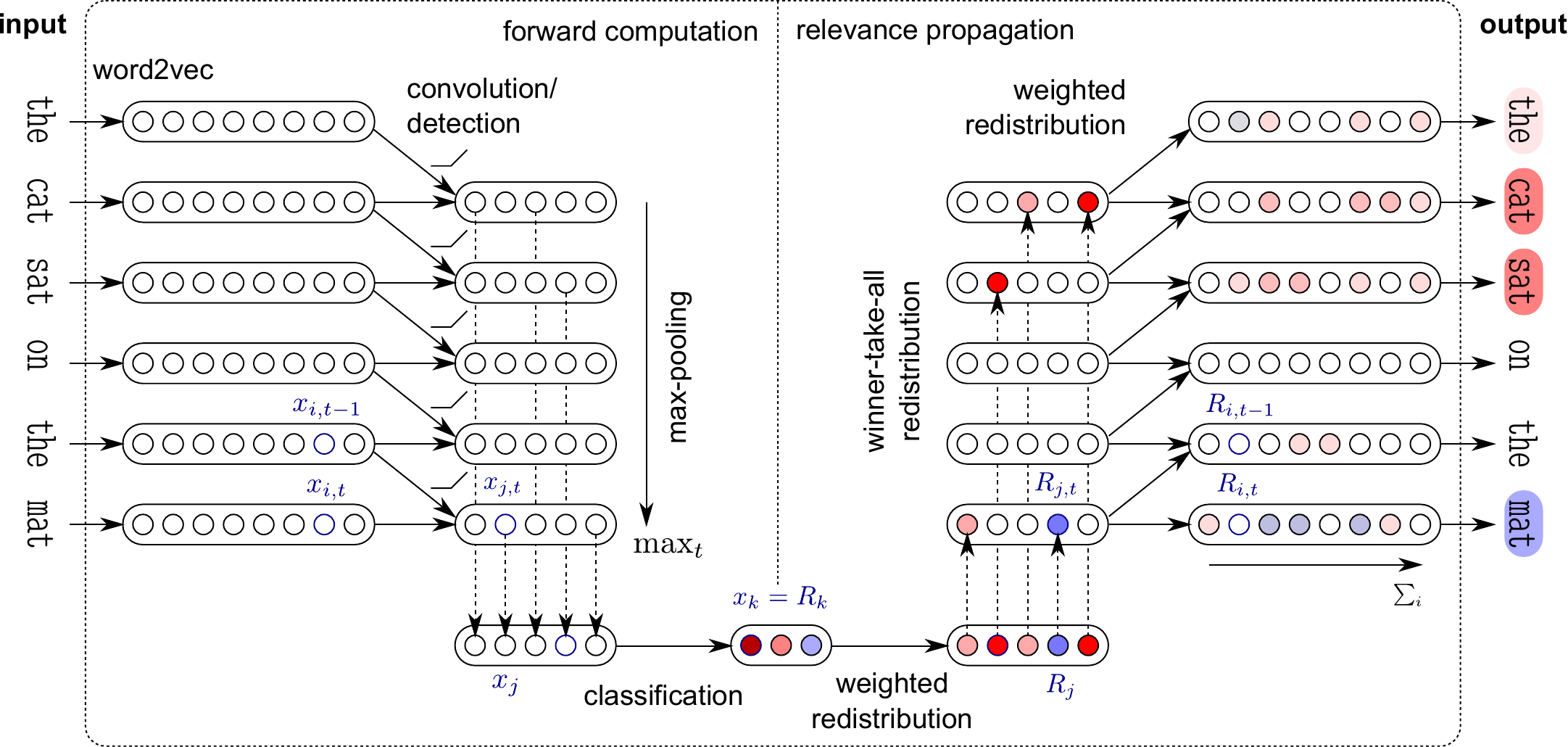}
\caption{Diagram of a CNN-based interpretable machine learning system consisting of a forward processing that computes for each input document a high-level concept (e.g.\ semantic category or sentiment), and a redistribution procedure that explains the prediction in terms of words.}
\label{fig:lrp}
\end{figure}

In this section we describe our method for identifying words in a text document, that are relevant with respect to a given category of a classification problem. For this, we assume that we are given a vector-based word representation and a neural network that has already been trained to map accurately documents to their actual category. Our method can be divided in four steps: (1) Compute an input representation of a text document based on word vectors. (2) Forward-propagate the input representation through the convolutional neural network until the output is reached. (3) Backward-propagate the output through the network using the layer-wise relevance propagation (LRP) method, until the input is reached. (4) Pool the relevance scores associated to each input variable of the network onto the words to which they belong. As a result of this four-step procedure, a decomposition of the prediction score for a category onto the words of the documents is obtained. Decomposed terms are called relevance scores. These relevance scores can be viewed as highlighted text or can be used to form a list of top-words in the document. The whole procedure is also described visually in Figure \ref{fig:lrp}. While we detail in this section the LRP method for a specific network architecture and with predefined choices of layers, the method can in principle be extended to any architecture composed of similar or larger number of layers.

At the end of this section we introduce different methods which will serve as baselines for comparison.
A baseline for the convolutional neural network model is the BoW/SVM classifier, with the LRP procedure adapted accordingly \cite{Bach}. 
A baseline for the LRP relevance decomposition procedure is gradient-based sensitivity analysis (SA), a technique which assigns sensitivity scores to individual words.
In the vector-based document representation experiments, we will also compare LRP to uniform and TFIDF baselines.

\subsection{Representing Words and Documents}

Prior to training the neural network and using it for prediction and explanation, we first derive a numerical representation of the text documents that will serve as an input to the neural classifier. To this end, we map each individual word in the document to a vector embedding, and concatenate these embeddings to form a matrix of size the number of words in the document times the dimension of the word embeddings.
A distributed representation of words can be learned from scratch, or fine-tuned simultaneously with the classification task of interest. In the present work, we use only pre-training as it was shown that, even without fine-tuning, this leads to good neural network classification performance for a variety of tasks like e.g.\ natural language tagging or sentiment analysis \cite{Collobert, Kim}.

One shallow neural network model for learning word embeddings from unlabeled text sources, is the continuous bag-of-words (CBOW) model of \cite{Mikolov1}, which is similar to the log-bilinear language model from \cite{Mnih2007, Mnih2012} but ignores the order of context words. In the CBOW model, the objective is to predict a target middle word from the average of the embeddings of the context words that are surrounding the middle word, by means of direct dot products between word embeddings. During training, a set of word embeddings for context words $v$ and for target words $v'$ are learned separately. After training is completed, only the context word embeddings $v$ will be retained for further applications. The CBOW objective has a simple maximum likelihood formulation, where one maximizes over the training data the sum of the logarithm of probabilities of the form:
$$
P (w_t |  w_{t-n:t+n} ) = \frac{\exp \Big( ( {1 \over{2n}}\cdot \; {\sum_{-n\leq j \leq n, j \neq 0}{\; v_{w_{t+j}}} )^\top v'_{w_t} \Big)}  }{\sum_{w \in V} \;  \exp \Big( ( {1 \over{2n}}\cdot \; {\sum_{-n\leq j \leq n, j \neq 0}{\; v_{w_{t+j}}})^\top v'_{w} \Big)}}
$$
where the softmax normalization runs over all words in the vocabulary $V$, $2n$ is the number of context words per training text window, $w_t$ represents the target word at the $t^\mathrm{th}$ position in the training data and $w_{t-n:t+n}$ represent the corresponding context words. 

In the present work, we utilize pre-trained word embeddings obtained with the CBOW architecture and the negative sampling training procedure \cite{Mikolov2}. We will refer to these embeddings as word2vec embeddings.

\subsection{Predicting Category with a Convolutional Neural Network}\label{sec3:predicting_category}

Our ML model for classifying text documents, is a word-embedding based convolutional neural network (CNN) model similar to the one proposed in \cite{Kim} for sentence classification, which itself is a slight variant of the model introduced in \cite{Collobert} for semantic role labeling. This architecture is depicted in Figure \ref{fig:lrp} (left) and is composed of several layers. 

As previously described, in a first step we map each word in the document to its word2vec vector. Denoting by $D$ the word embedding dimension and by $L$ the document length, our input is a matrix of shape $D \times L$. We denote by $x_{i,t}$ the value of the $i^\mathrm{th}$ component of the word2vec vector representing the  $t^\mathrm{th}$  word in the document. The convolution/detection layer produces a new representation composed of $F$ sequences indexed by $j$, where each element of the sequence is computed as:
$$
\forall{j,t}:~ x_{j,t} = {  \textstyle\max\Big(0, \; \sum_{i,\tau} x_{i,t-\tau} \; w^{(1)}_{i, j ,\tau} + b^{(1)}_j\Big) = \max\Big(0, \; \sum_{i} \; \big(x_{i} \ast w^{(1)}_{i,j}\big)_t + b^{(1)}_j\Big) }
$$
where $t$ indicates a position within the text sequence, $j$ designates a feature map, and $\tau \in \{0,1,\dots,H-1\}$ is a delay with range $H$  the filter size of the one-dimensional convolutional operation $\ast$. After the convolutional operation, which yields $F$ features maps of length $L-H+1$, we apply the ReLU non-linearity element-wise. Note that the trainable parameters $w^{(1)}$ and $b^{(1)}$ do not depend on the position $t$ in the text document, hence the convolutional processing is equivariant with this physical dimension. In Figure \ref{fig:lrp}, we use $\tau \in \{0,1\}$. 
The next layer computes, for each dimension $j$ of the previous representation, the maximum over the entire text sequence of the document:
$$
\forall{j}:~ x_j = {  \textstyle \max_t \big\{ x_{j,t} \big\} }
$$
This layer creates invariance to the position of the features in the document. 
Finally, the $F$ pooled features are fed into an endmost logistic classifier where the unnormalized log-probability of each of the $C$ classes, indexed by the variable $k$ are given by:
$$
\forall{k}:~ x_k = {  \textstyle { \sum_{j}} \;  x_j \; w^{(2)}_{jk} + b^{(2)}_k }
\label{eq:fclayer}
$$
where $w^{(2)}$, $b^{(2)}$ are trainable parameters of size $F \times C$ resp. size $C$ defining a fully-connected linear layer. The outputs $x_k$ can be converted to probabilities through the softmax function $p_k = \exp(x_k) / \sum_{k'} \exp(x_{k'})$.  For the LRP decomposition we take the unnormalized classification scores $x_k$ as a starting point. 

\subsection{Explaining Predictions with Layer-wise Relevance Propagation}\label{sec2:lrp}

Layer-wise relevance propagation (LRP) \cite{Bach, LapJMLR16} is a recently introduced technique for estimating which elements of a classifier input are important to achieve a certain classification decision.
It can be applied to bag-of-words SVM classifiers as well as to layer-wise structured neural networks.
For every input data point and possible target class, LRP delivers one scalar relevance value per input variable, hereby indicating whether the corresponding part of the input is contributing {\it for} or {\it against} a specific classifier decision, or if this input variable is rather uninvolved and irrelevant to the classification task at all.

The main idea behind LRP is to redistribute, for each possible target class separately, the output prediction score (i.e.\ a scalar value) that causes the classification, back to the input space via a backward propagation procedure that satisfies a layer-wise conservation principle.
Thereby each intermediate classifier layer up to the input layer gets allocated relevance values, and  the sum of the relevances per layer is equal to the classifier prediction score for the considered class. Denoting by $x_{i,t}\,, x_{j,t}\,, x_{j}\,, x_{k}$ the neurons of the CNN layers presented in the previous section, we associate to each of them respectively a relevance score $R_{i,t}\,, R_{j,t}\,, R_j\,, R_k$. Accordingly the layer-wise conservation principle can be written as:
\begin{equation}
{\textstyle \sum_{i,t} R_{i,t} = \sum_{j,t} R_{j,t} = \sum_j R_j = \sum_k R_k}
\end{equation}
where each sum runs over all neurons of a given layer of the network.
To formalize the redistribution process from one layer to another, we introduce the concept of messages $R_{a \leftarrow b}$ indicating how much relevance circulates from a given neuron $b$ to a neuron $a$ in the next lower-layer. We can then express the relevance of neuron $a$ as a sum of incoming messages using: ${ \textstyle R_a = \sum_{b \in {\text{upper}(a)}} R_{a \leftarrow b}}$ where ${\text{upper}(a)}$ denotes the upper-layer neurons connected to $a$.
To bootstrap the propagation algorithm, we set the top-layer relevance vector to $\forall_k: R_k = x_k \cdot \delta_{kc}$ where $\delta$ is the Kronecker delta function, and $c$ is the target class of interest for which we would like to explain the model prediction in isolation from other classes. 

In the top fully-connected layer, messages are computed following a weighted redistribution formula:
\begin{equation}\label{eq:rel-fclayer}
R_{j \leftarrow k} = \frac{z_{jk}}{\sum_{j} z_{jk}} R_k
\end{equation}
where we define $z_{jk} = x_j w^{(2)}_{jk} + F^{-1} (b^{(2)}_k + \epsilon \cdot (1_{x_k \geq 0} - 1_{x_k < 0}))$. This formula redistributes relevance onto lower-layer neurons in proportions to $z_{jk}$ representing the contribution of each neuron to the upper-layer neuron value in the forward propagation, incremented with a small stabilizing term $\epsilon$ that prevents the denominator from nearing zero, and hence avoids too large positive or negative relevance messages. In the limit case where  $\epsilon \rightarrow \infty$, the relevance is redistributed uniformly along the network connections. As a stabilizer value we  use $\epsilon = 0.01$ as introduced in \cite{Bach}. After computation of the messages according to Equation \ref{eq:rel-fclayer}, the latter can be pooled onto the corresponding neuron by the formula $R_j = \sum_k R_{j \leftarrow k}$.

The relevance scores $R_j$ are then propagated through the max-pooling layer using the formula:
\begin{equation}\label{eq:rel-maxlayer}
R_{j,t} = \left\{
\begin{array}{ll}
R_j & \text{if} \; \;  t = \mathrm{arg}\max_{t'} \; x_{j,t'}\\
0 & \text{else}
\end{array}
\right.
\end{equation}
which is a ``winner-take-all'' redistribution analogous to the rule used during training for backpropagating gradients, i.e.\ the neuron that had the maximum value in the pool is granted all the relevance from the upper-layer neuron. 
Finally, for the convolutional layer we use the weighted redistribution formula:
\begin{equation}\label{eq:rel-convlayer}
R_{(i,t-\tau) \leftarrow (j,t)} = \frac{z_{i, j, \tau}}{ \sum_{i,\tau} z_{i, j, \tau}}
\end{equation}
where $z_{i, j, \tau} = x_{i,t-\tau} w^{(1)}_{i, j, \tau} + (HD)^{-1} (b^{(1)}_j + \epsilon \cdot (1_{x_{j,t} > 0} - 1_{x_{j,t} \leq 0}))$, which is similar to Equation \ref{eq:rel-fclayer} except for the increased notational complexity incurred by the convolutional structure of the layer. Messages can finally be pooled onto the input neurons by computing $R_{i,t} = \sum_{j,\tau} R_{(i,t) \leftarrow (j,t+\tau)}$.

\subsection{Word Relevance and Vector-Based Document Representation}\label{sec3:pooling}

So far, the relevance has been redistributed only onto individual components of the word2vec vector associated to each word, in the form of single input neuron relevances $R_{i,t}$. To obtain a word-level relevance value, one can pool the relevances over all dimensions of the word2vec vector, that is compute:
\begin{equation}
R_t = {\textstyle \sum_i} R_{i,t}
\end{equation}
and use this value to highlight words in a text document, as shown in Figure \ref{fig:lrp} (right). These word-level relevance scores can further be used to condense the semantic information of text documents, by building vectors $\boldsymbol{d} \in \mathbb{R}^D$ representing full documents through linearly combining word2vec vectors:
\begin{equation}\label{eq:summary_vector_cnn_word_level}
\forall_i:~d_i = {\textstyle \sum_t} \; R_{t} \cdot x_{i,t}
\end{equation}
The vector $\boldsymbol{d}$ is a summary that consists of an additive composition of the semantic representation of all relevant words in the document. Note that the resulting document vector lies in the same semantic space as word2vec vectors. A more fined-grained extraction technique does not apply word-level pooling as an intermediate step and extracts only the relevant subspace of each word:
\begin{equation}\label{eq:summary_vector_cnn_element_wise}
\forall_i:~d_i = {\textstyle \sum_t} \; R_{i,t} \cdot x_{i,t}
\end{equation}
This last approach is particularly useful to address the problem of word homonymy, and will thus result in even finer semantic extraction from the document.
In the remaining we will refer to the semantic extraction defined by Eq.~\ref{eq:summary_vector_cnn_word_level} as word-level extraction, and to the one from Eq.~\ref{eq:summary_vector_cnn_element_wise} as element-wise (ew) extraction. In both cases we call vector $\boldsymbol{d}$ a {\it document summary vector}.

\subsection{Baseline Methods}\label{sec3:alternative_methods}

In the following we briefly mention methods which will serve as baselines for comparison.

\vspace*{0.25cm}
\noindent{\bf Sensitivity Analysis.} Sensitivity analysis (SA) \cite{Dimopoulos95, Gevrey03, BaehrensSHKHM10} assigns scores $R_{i,t} = (\partial x_k / \partial x_{i,t})^2$ to input variables representing  the steepness of the decision function in the input space. These partial derivatives are straightforward to compute using standard gradient propagation \cite{rumelhart86} and are readily available in most neural network implementations. Hereby we note that sensitivity analysis redistributes the quantity $\|\nabla x_k\|{_2^2}$, while LRP redistributes $x_k$. However, the local steepness information is a relatively weak proxy of the actual function value, which is the real quantity of interest when estimating the contribution of input variables w.r.t.\ to a current classifier's decision. We further note that relevance scores obtained with LRP are signed, while those obtained with SA are positive.

\vspace*{0.25cm}
\noindent{\bf BoW/SVM.} As a baseline to the CNN model, a bag-of-words linear SVM classifier will be used to predict the document categories. In this model each text document is first mapped to a vector $x$ with dimensionality $V$ the size of the training data vocabulary, where each entry is computed as a term frequency - inverse document frequency (TFIDF) score of the corresponding word. Subsequently these vectors $x$ are normalized to unit euclidean norm. In a second step, using the vector representations $x$ of all documents, $C$ maximum margin separating hyperplanes are learned to separate each of the classes of the classification problem from the other ones.
As a result we obtain for each class $c \in C$ a linear prediction score of the form $s_c = w_c^\top x + b_c$, where $w_c\in \mathbb{R}^{V} $ and $b_c \in \mathbb{R}$ are class specific weights and bias. In order to obtain a LRP decomposition of the prediction score $s_c$ for class $c$ onto the input variables, we simply compute $ \textstyle R_i = (w_c)_i \cdot x_i + {b_c / D} $, where $D$ is the number of non-zero entries of $x$. Respectively, the sensitivity analysis redistribution of the prediction score squared gradient reduces to $R_i = (w_c)_i^2$.

Note that the BoW/SVM model being a linear predictor relying directly on word frequency statistics, it lacks expressive power in comparison to the CNN model which additionally learns intermediate hidden layer representations and convolutional filters. Moreover the CNN model can take advantage of the semantic similarity encoded in the distributed word2vec representations, while for the BoW/SVM model all words are ``equidistant'' in the bag-of-words semantic space. As our experiments will show, these limitations lead the BoW/SVM model to sometimes identify spurious words as relevant for the classification task.
In analogy to the semantic extraction proposed in section \ref{sec3:pooling} for the CNN model, we can build vectors $\boldsymbol{d}$ representing documents by leveraging the word relevances obtained with the BoW/SVM model. To this end, we introduce a binary vector $\tilde{x} \in \mathbb{R}^{V} $ whose entries are equal to one when the corresponding word from the vocabulary is present in the document and zero otherwise (i.e.\ $\tilde{x}$ is a binary bag-of-words representation of the document). Thereafter, we build the document summary vector $\boldsymbol{d}$ component-wise, so that $\boldsymbol{d}$ is just a vector of word relevances:
\begin{equation}\label{eq:summary_vector_svm}
\forall_i:~d_i =  R_{i} \cdot {\tilde{x}}_{i}
\end{equation}

\noindent{\bf Uniform/TFIDF based Document Summary Vector.} In place of the word-level relevance $R_t$ resp.\ $R_i$ in Eq.~\ref{eq:summary_vector_cnn_word_level} and Eq.~\ref{eq:summary_vector_svm}, we can use a uniform weighting. This corresponds to build the document vector $\boldsymbol d$ as an average of word2vec word embeddings in the first case, and to take as a document representation $\boldsymbol d$ a binary bag-of-words vector in the second case.
Moreover, we can replace $R_t$ in Eq.~\ref{eq:summary_vector_cnn_word_level} by an inverse document frequency (IDF) score, and $R_i$ in Eq.~\ref{eq:summary_vector_svm} by a TFIDF score. Both correspond to TFIDF weighting of either word2vec vectors, or of one-hot vectors representing words.

\section{Quality of Word Relevances and Model Explanatory Power}\label{sec4}
In this section we describe how to evaluate and compare the outcomes of algorithms which assign relevance scores to words (such as LRP or SA) through intrinsic validation.
Furthermore, we propose a measure of model explanatory power based on an extrinsic validation procedure.
The latter will be used to analyze and compare the relevance decompositions or {\it explanations} obtained with the neural network and the BoW/SVM classifier. Both types of evaluations will be carried out in section \ref{sec5}.

\subsection{Measuring the Quality of Word Relevances through Intrinsic Validation}

An evaluation of how good a method identifies relevant words in text documents can be performed qualitatively, e.g. at the document level, by inspecting the heatmap visualization of a document, or by reviewing the list of the most (or of the least) relevant words per document. A similar analysis can also be conducted at the dataset level, e.g. by compiling the list of the most relevant words for one category across all documents. The latter allows one to identify words that are representatives for a document category, and eventually to detect potential dataset biases or classifier specific drawbacks. 
However, in order to quantitatively compare algorithms such as LRP and SA regarding the identification of relevant words, we need an objective measure of the quality of the explanations delivered by relevance decomposition methods.
To this end we adopt an idea from \cite{SamArXiv15}: A word $w$ is considered highly relevant for the classification $f(x)$ of the document $x$ if removing it and classifying the modified document $\tilde{x}$ results in a strong decrease of the classification score $f(\tilde{x})$. This idea can be extended by sequentially deleting words from the most relevant to the least relevant or the other way round. The result is a graph of the prediction scores $f(\tilde{x})$ as a function of the number of deleted words. 
In our experiments, we employ this approach to track the changes in classification performance when successively deleting words according to their relevance value. By comparing the relative impact on the classification performance induced by different relevance decomposition methods, we can estimate how appropriate these methods are at identifying words that are really important for the classification task at hand.
The above described procedure constitutes an intrinsic validation, as it does not rely on an external classifier.

\subsection{Measuring Model Explanatory Power through Extrinsic Validation}\label{sec4:extrinsic_validation}

Although intrinsic validation can be used to compare relevance decomposition methods for a given ML model, this approach is not suited to compare the explanatory power of different ML models, since the latter requires a common evaluation basis.
Furthermore, even if we would track the classification performance changes induced by different ML models using an external classifier, it would not necessarily increase comparability, because removing words from a document may affect different classifiers very differently, so that their graphs $f(\tilde{x})$ are not comparable. 
Therefore, we propose a novel measure of model explanatory power which does not depend on a classification performance change, but only on the word relevances.
Hereby we consider ML model A as being more explainable than ML model B if its word relevances are more ``semantic extractive'', i.e. more helpful for solving a semantic related task such as the classification of document summary vectors.

More precisely, in order to quantify the  ML model explanatory power we undertake the following steps:\\[+3px]
(1) Compute document summary vectors for all test set documents using Eq.~\ref{eq:summary_vector_cnn_word_level}~or~\ref{eq:summary_vector_cnn_element_wise} for the CNN and  Eq.~\ref{eq:summary_vector_svm} for the BoW/SVM model.
Hereby use the ML model's predicted class as target class for the relevance decomposition (i.e. the summary vector generation is unsupervised).\\[+3px]
(2) Normalize the document summary vectors to unit euclidean norm, and perform a K-nearest-neighbors (KNN) classification of half of these vectors, using the other half of summary vectors as neighbors (hereby use standard KNN classification, i.e. nearest neighbors are identified by euclidean distance and neighbor votes are weighted uniformly). Use different hyperparameters $K$. \\[+3px]
(3) Repeat step (2) over 10 random data splits, and average the KNN classification accuracies  for each $K$. Finally, report the maximum (over different $K$) KNN accuracy as explanatory power index (EPI). The higher this value, the more {\it explanatory power} the ML model and the corresponding document summary vectors, will have.

In a nutshell, our EPI metric of explanatory power of a given ML model ``$f$'', combined with a relevance map ``$R$'', can informally be summarized as:
\begin{align}
\boldsymbol{d}(x) &= {\textstyle \sum_t} \; [R (f (x)) \odot   x]_t \nonumber\\[2mm]
{\text{EPI}}(f,R) \; &=  \; \max_{K} \; \; \texttt{KNN\_accuracy} \Big(\{\boldsymbol{d}(x^{(1)}),\dots,\boldsymbol{d}(x^{(N)})\},K\Big)
\label{eq:explain}
\end{align}
where $\boldsymbol{d}(x)$ is the document summary vector for input document $x$, and subscript $t$ denotes the words in the document. Thereby the sum $\sum_t$ and element-wise multiplication $\odot$ operations stand for the weighted combination specified explicitly in Eq.~\ref{eq:summary_vector_cnn_word_level}~-~\ref{eq:summary_vector_svm}. The KNN accuracy is estimated over all test set document summary vectors indexed from $1$ to $N$, and $K$ is the number of neighbors.

In the proposed evaluation procedure, the use of KNN as a common external classifier enables us to unbiasedly and objectively compare different ML models, in terms of the density and local neighborhood structure of the semantic information
extracted via the summary vectors in input feature space. Indeed we recall that summary vectors constructed via Eq.~\ref{eq:summary_vector_cnn_word_level}~and~\ref{eq:summary_vector_cnn_element_wise} lie in the same semantic space as word2vec embeddings, and that summary vectors obtained via Eq.~\ref{eq:summary_vector_svm} live in the bag-of-words space.

\section{Results}\label{sec5}
This section summarizes our experimental results.
We first describe the dataset, experimental setup, training procedure and classification accuracy of our ML models. We will consider four ML models: three CNNs with different filter sizes and a BoW/SVM classifier.
Then, we demonstrate that LRP can be used to identify relevant words in text documents.
We compare heatmaps for the best performing CNN model and the BoW/SVM classifier, and report the most representative words for three exemplary document categories.
These results demonstrate qualitatively that the CNN model produces better explanations than the BoW/SVM classifier.
After that we move to the evaluation of the document summary vectors, where we show that a 2D PCA projection of the document vectors computed from the LRP scores
groups documents according to their topics (without requiring the true labels). Since worse results are obtained when using the SA scores or the uniform or TFIDF weighting, this indicates that the explanations produced by LRP are semantically more meaningful than the latter. Finally, we confirm quantitatively the observations made before, namely that (1) the LRP decomposition method provides better explanations than SA and that (2) the CNN model outperforms the BoW/SVM classifier in terms of explanatory power.

\subsection{Experimental Setup}\label{20news_setup}

\subsubsection{Dataset}
For our experiments we consider a topic categorization task, and employ the freely available 20Newsgroups\footnote{20Newsgroups dataset available at \texttt{http://qwone.com/\%7Ejason/20Newsgroups/}} dataset consisting of newsgroup posts evenly distributed among twenty fine-grained categories.
More precisely we use the 20news-bydate version, which is already partitioned into 11314 training and 7532 test documents corresponding to different periods in time.

\subsubsection{Preprocessing and Training}
As a first preprocessing step, we remove the headers from the documents (by splitting at the first blank line) and tokenize the text with NLTK\footnote{Natural Language Toolkit available at \texttt{http://www.nltk.org} (tokenizer  \textit{sent\_tokenize} and \textit{word\_tokenize}, module \textit{nltk.tokenize})}. Then, we filter the tokenized data by retaining only tokens composed of the following four types of characters: alphabetic, hyphen, dot and apostrophe, and containing at least one alphabetic character. Hereby we aim to remove punctuation, numbers or dates, while keeping abbreviations and compound words. We do not apply any further preprocessing, as for instance stop-word removal or stemming, except for the SVM classifier where we additionally perform lowercasing, as this is a common setup for bag-of-words models.
We truncate the resulting sequence of tokens to a chosen fixed length of 400 in order to simplify neural network training (in practice our CNN can process any arbitrary sized document).
Lastly, we build the neural network input by horizontally concatenating pre-trained word embeddings, according to the sequence of tokens appearing in the preprocessed document. In particular, we take the $300$-dimensional freely available\footnote{word2vec embeddings available at \texttt{https://code.google.com/p/word2vec/}} word2vec embeddings \cite{Mikolov2}. Out-of-vocabulary words are simply initialized to zero vectors. As input normalization, we subtract the mean and divide by the standard deviation obtained over the flattened training data. We train the neural network by minimizing the cross-entropy loss via mini-batch stochastic gradient descent using $l_2$-norm and dropout as regularization. We tune the ML model hyperparameters by 10-fold cross-validation in case of the SVM, and by employing 1000 random documents as fixed validation set for the CNN model. However, for the CNN hyperparameters we did not perform an extensive grid search and stopped the tuning once we obtained models with reasonable classification performance for the purpose of our experiments.


Table~\ref{tab:20news_accuracy} summarizes the performance of our trained models. Herein CNN1, CNN2, CNN3 respectively denote neural networks with convolutional filter size $H$ equal to 1, 2 and 3 (i.e.\ covering 1, 2 or 3 consecutive words in the document). One can see that the linear SVM performs on par with the neural networks, i.e.\ the non-linear structure of the CNN models does not yield a considerable advantage toward classification accuracy. Similar results have also been reported in previous studies \cite{Zhang}, where it was observed that for document classification a convolutional neural network model starts to outperform a TFIDF-based linear classifier only on datasets in the order of millions of documents. This can be explained by the fact that for most topic categorization tasks, the different categories can be separated linearly in the very high-dimensional bag-of-words or bag-of-N-grams space thanks to sufficiently disjoint sets of features. 

\begin{table}[!h]
\centering\small
\caption{Test set performance of the ML models for 20-class document classification.}\label{tab:20news_accuracy}
\begin{tabular}{lcc}
\toprule
ML Model 					&      Test Accuracy (\%) 	&      \\  
\midrule
BoW/SVM  ($V= 70631$ words)			&      80.10 			
\\
CNN1 ($H=1$, $F=600$)				&      79.79 			
\\
CNN2 ($H=2$, $F=800$) 				&  \bf 80.19 			
\\
CNN3 ($H=3$, $F=600$) 				&      79.75 			
\\
\bottomrule
\end{tabular}
\end{table}


\subsection{Identifying Relevant Words}
\label{section:relevantwords}
\begin{figure}
\centering
\boxed{\includegraphics[width=1.0\textwidth, height=7.5cm,clip=True,trim=60 232 75 40]{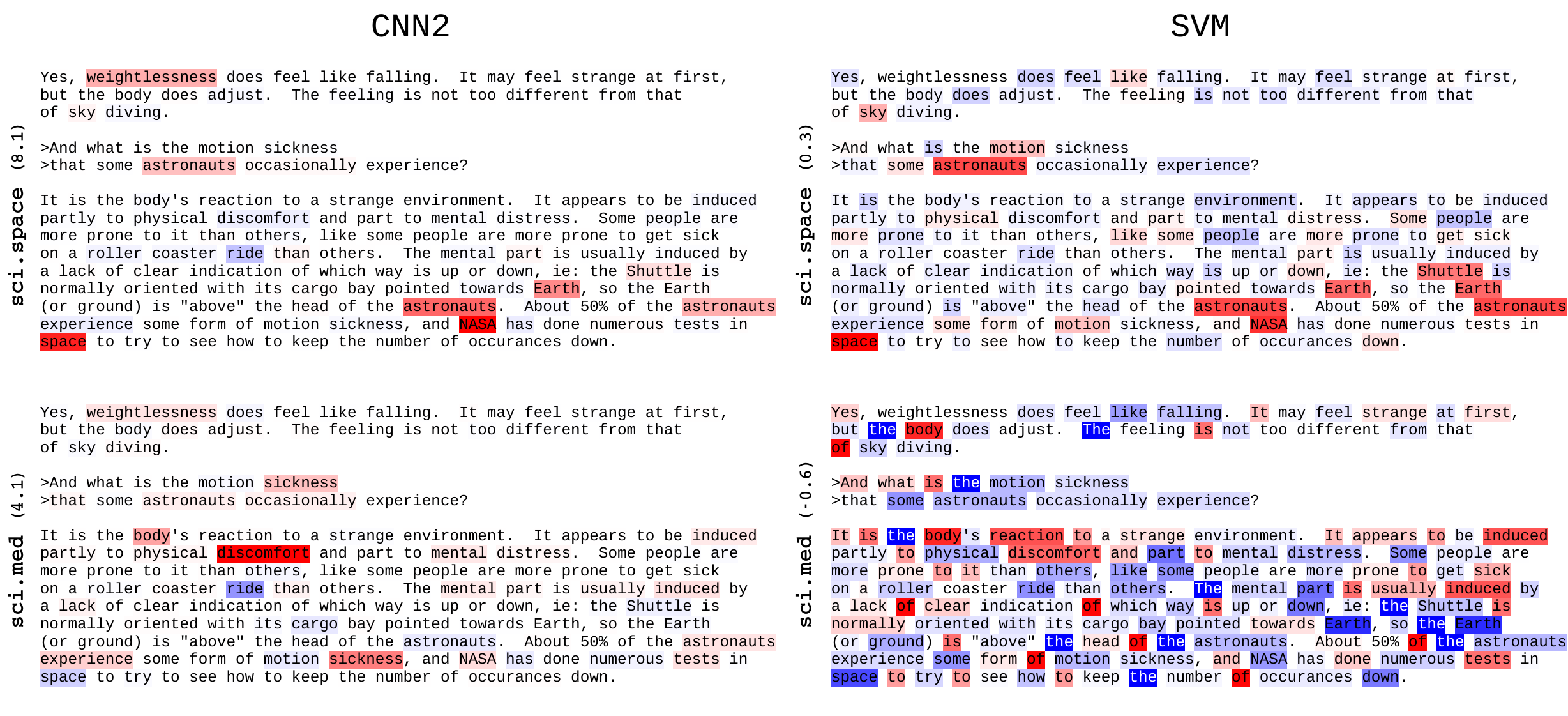}}
\caption{LRP heatmaps of the document \texttt{sci.space 61393} for the CNN2 and SVM model.
Positive relevance is mapped to red, negative to blue. The color opacity is normalized to the maximum absolute relevance per document. The LRP target class and corresponding classification prediction score is indicated on the left.}\label{fig:20news_heatmap}

\centering \small
\includegraphics[width=1\textwidth]{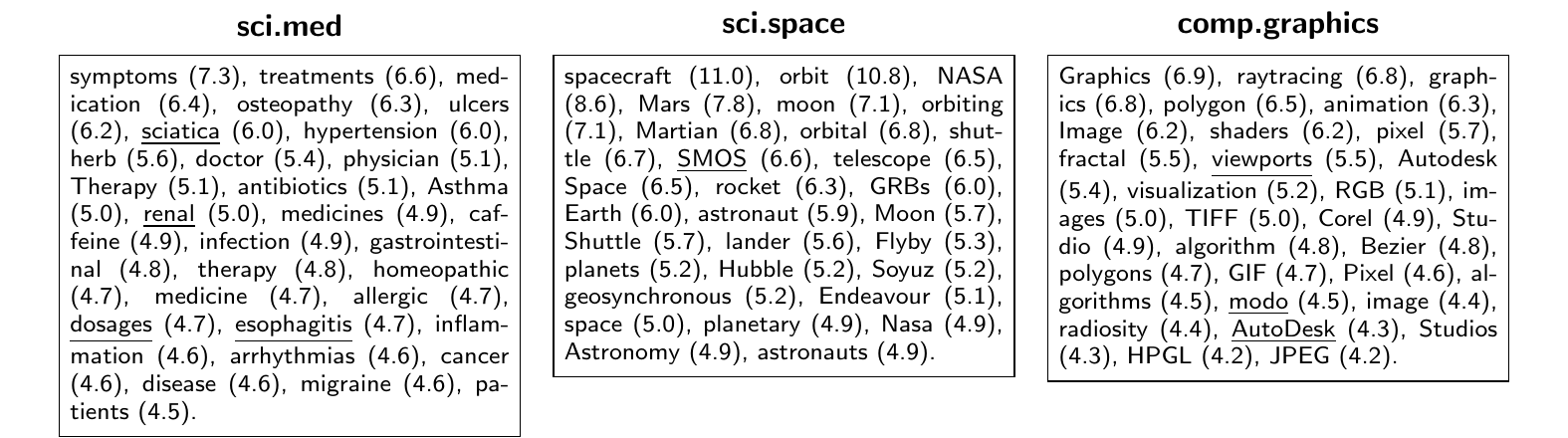}
\caption{The 30 most relevant words per class for the CNN2 model listed in decreasing order of their relevance (value indicated in parentheses). Underlined words do not occur in the training data.}\label{fig:words_cnn2}

\centering \small
\includegraphics[width=1\textwidth]{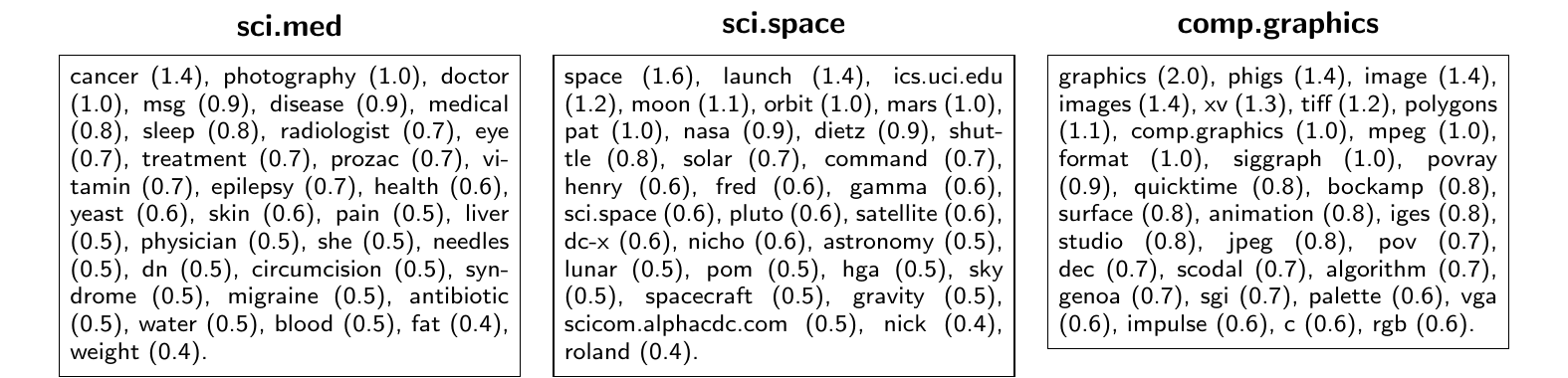}
\caption{The 30 most relevant words per class for the BoW/SVM model listed in decreasing order of their relevance (value indicated in parentheses). Underlined words do not occur in the training data.}
\label{fig:words_svm}
\end{figure}

Figure~\ref{fig:20news_heatmap} compiles the resulting LRP heatmaps we obtain on an exemplary \texttt{sci.space} test document that is correctly classified by the SVM and the best performing neural network model CNN2.
Note that for the SVM model the relevance values are computed per bag-of-words feature, i.e., same words will have same relevance irrespectively of their context in the document, whereas for the CNN classifier we visualize one relevance value per word position.
Hereby we consider as target class for the LRP decomposition the classes \texttt{sci.space} and \texttt{sci.med}. We can observe that the SVM model considers insignificant words like \textit{the}, \textit{is}, \textit{of} as very relevant (either negatively or positively) for the target class \texttt{sci.med}, and at the same time mistakenly estimates words like \textit{sickness}, \textit{mental} or \textit{distress} as negatively contributing to this class (indicated by blue coloring), while on the other hand the CNN2 heatmap is consistently more sparse and concentrated on semantically meaningful words. 
This sparsity property can be attributed to the max-pooling non-linearity which for each feature map in the neural network selects the first most relevant feature that occurs in the document. 
As can be seen, it significantly simplifies the interpretability of the results by a human.
Another disadvantage of the SVM model is that it relies entirely on local and global word statistics, thus can only assign relevances proportionally to the TFIDF BoW features (plus a class-dependent bias term), while the neural network model benefits from the knowledge encoded in the word2vec embeddings.    
For instance, the word {\it weightlessness} is not highlighted by the SVM model for the target class \texttt{sci.space}, because this word does not occur in the training data and thus is simply ignored by the SVM classifier. The neural network however is able to detect and attribute relevance to unseen words thanks to the semantical information encoded in the pre-trained word2vec embeddings.

As a dataset-wide analysis, we determine the words identified through LRP as constituting class representatives. For that purpose we set one class as target class for the relevance decomposition, and conduct LRP over all test set documents (i.e.\ irrespectively of the true or ML model's predicted class). Subsequently, we sort all the words appearing in the test data in decreasing order of the obtained word-level relevance values, and retrieve the thirty most relevant ones. The result is a list of words identified via LRP as being highly supportive for a classifier decision toward the considered class. Figures~\ref{fig:words_cnn2} and \ref{fig:words_svm} list the most relevant words for different LRP target classes, as well as the corresponding word-level relevance values for the CNN2 and the SVM model.
Through underlining we indicate words that do not occur in the training data. 
Interestingly, we observe that some of the most ``class-characteristical'' words identified via the neural network model correspond to words that do not even appear in the training data. In contrast, such words are simply ignored by the SVM model as they do not occur in the bag-of-words vocabulary. 
Similarly to the previous heatmap visualizations, the class-specific analysis reveals that the SVM classifier occasionally assigns high relevances to semantically insignificant words like for example the pronoun \textit{she} for the target class \texttt{sci.med} (20$^{th}$ position in left column of Fig.~\ref{fig:words_svm}), or to the names \textit{pat}, \textit{henry}, \textit{nicho} for the target the class \texttt{sci.space} (resp.\  7, 13, 20$^{th}$ position in middle column of Fig.~\ref{fig:words_svm}). In the former case the high relevance is due to a high term frequency of the word (indeed the word {\it she} achieves its highest term frequency in one \texttt{sci.med} test document where it occurs 18 times), whereas in the latter case this can be explained by a high inverse document frequency or by a class-biased occurrence of the corresponding word in the training data ({\it pat} appears within 16 different training document categories but 54.1\% of its occurrences are within the category \texttt{sci.space} alone, 79.1\% of the 201 occurrences of {\it henry} appear among \texttt{sci.space} training documents, and {\it nicho} appears exclusively in nine \texttt{sci.space} training documents). On the contrary, the neural network model seems less affected by word counts regularities and systematically attributes the highest relevances to words semantically related to the considered target class.
These results demonstrate that, subjectively, the neural network is better suited to identify relevant words in text documents than the BoW/SVM model.

\subsection{Document Summary Vectors}
\label{section:summaryvectors}
The word2vec embeddings are known to exhibit linear regularities representing semantical relationships between words \cite{Mikolov1,Mikolov2}.
We explore whether these regularities can be transferred to a new document representation, which we denoted as document summary vector, when building this vector as a weighted combination of word2vec embeddings (see Eq.~\ref{eq:summary_vector_cnn_word_level} and Eq.~\ref{eq:summary_vector_cnn_element_wise}) or as a combination of one-hot word vectors (see Eq.~\ref{eq:summary_vector_svm}).
We compare the weighting scheme based on the LRP relevances to the following baselines: SA relevance, TFIDF and uniform weighting (see section \ref{sec3:alternative_methods}).

\begin{figure}[!h]
\centering
\includegraphics[width=0.8\textwidth]{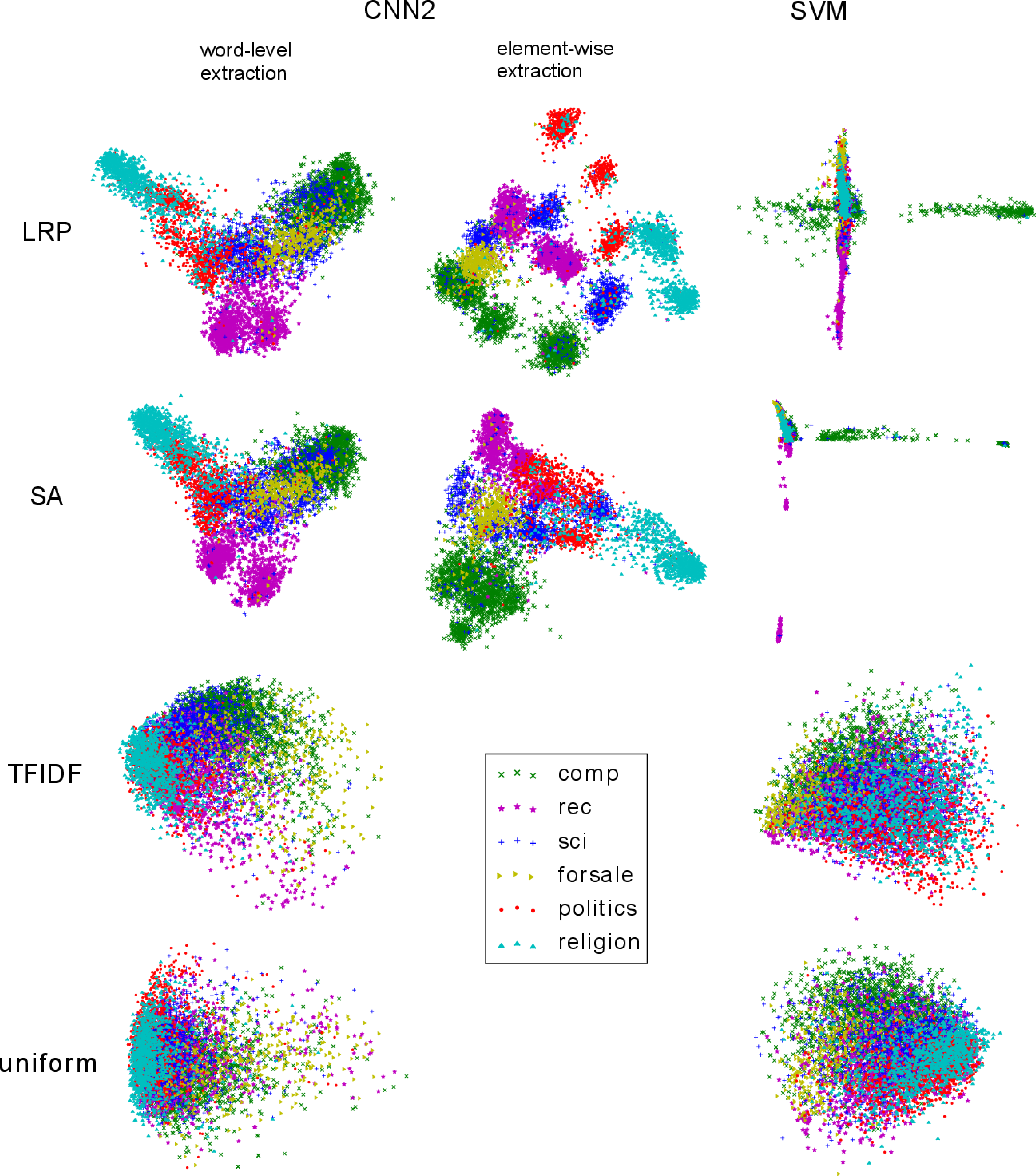}
\caption{PCA projection of the summary vectors of the 20Newsgroups test documents. The LRP/SA based weightings were computed using the ML model's predicted class, the colors denote the true labels.}\label{fig:20news_pca_CNN}
\end{figure}

The two-dimensional PCA projection of the summary vectors obtained via the CNN2 resp.\ the SVM model, as well as the corresponding TFIDF/uniform weighting baselines are shown in Figure~\ref{fig:20news_pca_CNN}. In these visualizations we group the 20Newsgroups test documents into six top-level categories (the grouping is performed according to the dataset website), and we color each document according to its {\it true} category (note however that, as mentioned earlier, the relevance decomposition is always performed in an unsupervised way, i.e., with the ML model's {\it predicted} class). 
For the CNN2 model, we observe that the two-dimensional PCA projection reveals a clear-cut clustered structure when using the element-wise LRP weighting for semantic extraction, while no such regularity is observed with uniform or TFIDF weighting. The word-level LRP or SA weightings, as well as the element-wise SA weighting present also a form of bundled layout, but not as dense and well-separated as in the case of element-wise LRP.
For the SVM model, the two-dimensional visualization of the summary vectors exhibits partly a cross-shaped layout for LRP and SA weighting, while again no particular structure is observed for TFIDF or uniform semantic extraction. This analysis confirms the observations made in the last section, namely that the neural network outperforms the BoW/SVM classifier in terms of explainability.
Figure~\ref{fig:20news_pca_CNN} furthermore suggests that LRP provides semantically more meaningful semantic extraction than the baseline methods.
In the next section we will confirm these observations quantitatively.

\subsection{Quantitative Evaluation}
\subsubsection{How good does LRP identify relevant words ?}
\label{section:worddeletion}
In order to quantitatively validate the hypothesis that LRP is able to identify words that either {\it support} or {\it inhibit} a specific classifier decision, we conduct several word-deleting experiments on the CNN models using LRP scores as relevance indicator. More specifically, in accordance to the word-level relevances we delete a sequence of words from each document, re-classify the documents with ``missing words'', and report the classification accuracy as a function of the number of deleted words. Hereby the word-level relevances are computed on the original documents (with no words deleted).
For the deleting experiments, we consider only 20Newsgroups test documents that have a length greater or equal to 100 tokens (after prepocessing), this amounts to 4963 test documents, from which we delete up to 50 words. 
For deleting a word we simply set the corresponding word embedding to zero in the CNN input.
Moreover, in order to assess the pertinence of the LRP decomposition method as opposed to alternative relevance models, we additionally perform word deletions according to SA word relevances, as well as random deletion. In the latter case we sample a random sequence of 50 words per document, and delete the corresponding words successively from each document. We repeat the random sampling 10 times, and report the average results (the standard deviation of the accuracy is less than 0.0141 in all our experiments). We additionally perform a biased random deletion, where we sample only among words comprised in the word2vec vocabulary (this way we avoid to delete words we have already initialized as zero-vectors as there are out of the word2vec vocabulary, however as our results show this biased deletion is almost equivalent to strict random selection).
 
\begin{figure}[!h]
\centering
\includegraphics[width=0.95\textwidth]{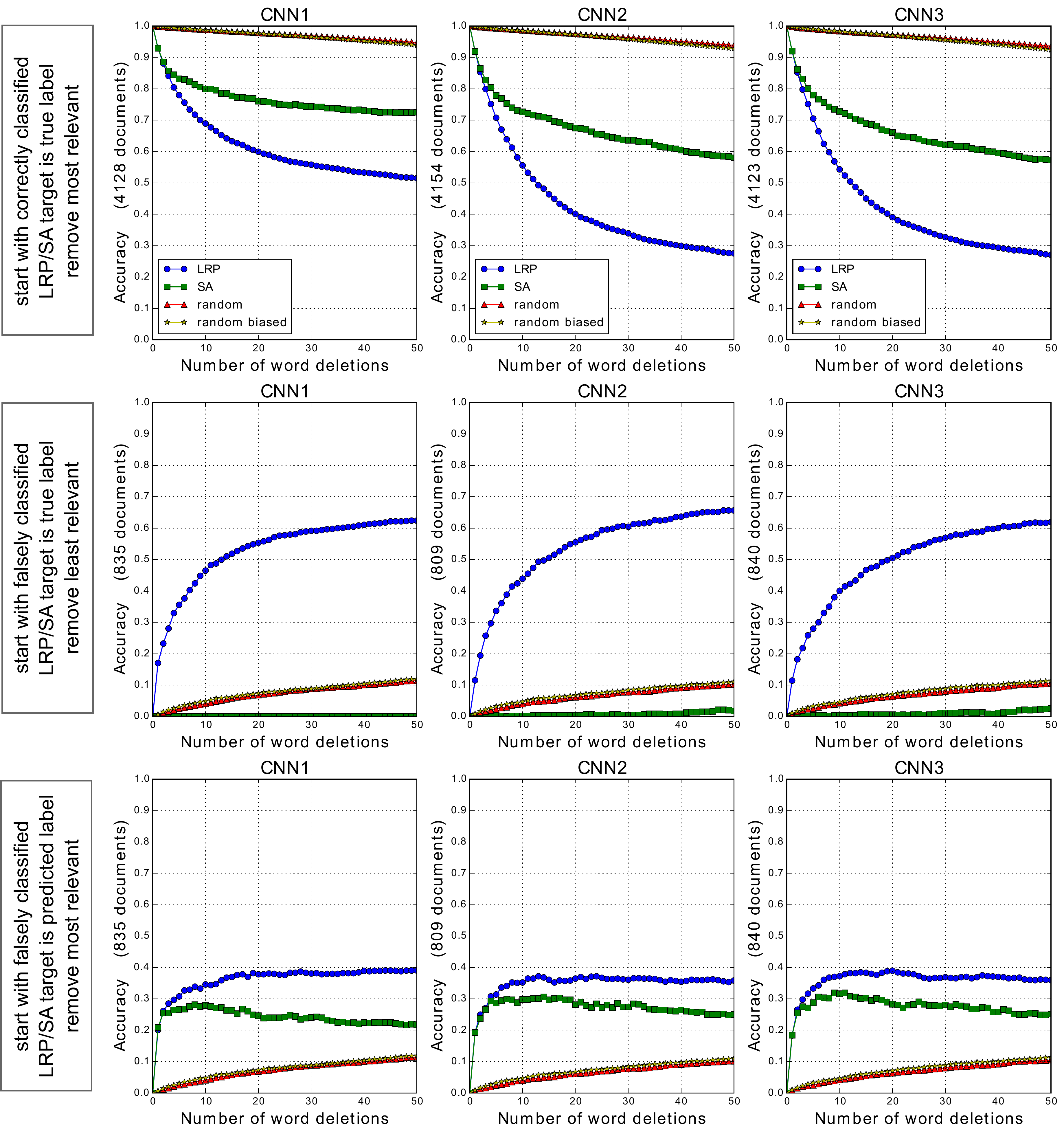}
\caption{Word deletion experiments for the CNN1, CNN2 and CNN3 model. The LRP/SA target class is either the true document class, and words are deleted in decreasing (first row, lower curve is better) resp.\ increasing (second row, higher curve is better) order of their LRP/SA relevance, or else the target class is the predicted class (third row, higher curve is better) in which case words are deleted in decreasing order of their relevance. Random (biased) deletion is reported as average over 10 runs.}\label{fig:deletion}
\end{figure} 
 
As a first deletion experiment, we start with the subset of test documents that are initially correctly classified by the CNN models, and  successively delete words in decreasing order of their LRP/SA word-level relevance. In this first deletion experiment, the LRP/SA relevances are computed with the true document class as target class for the relevance decomposition.
In a second experiment, we perform the opposite evaluation. Here we start with the subset of initially falsely classified documents, and delete successively words in increasing order of their relevance, while considering likewise the true document class as target class for the relevance computation.
In the third experiment, we start again with the set of initially falsely classified documents, but now delete words in decreasing order of their relevance, considering the classifier's initially predicted class as target class for the relevance decomposition.

Figure~\ref{fig:deletion} summarizes the resulting accuracies when deleting words resp.\ from the CNN1, CNN2 and CNN3 input documents (each row in the figure corresponds to one of the three deletion experiments). Note that we do not report results for the BoW/SVM model, as our focus here is the comparison between LRP and SA and not between different ML models\footnote{Besides we note that intrinsic validation is also not the right tool for comparing the BoW/SVM and the CNN models, as the resulting accuracies are not directly comparable (deleting a word from the bag-of-words document representation has a different effect than setting a word to zero in the CNN input).}.
Through successive deleting of either ``positive-relevant'' words in decreasing order of their LRP relevance, or of ``negative-relevant'' words in increasing order of their LRP relevance, we confirm that both extremal LRP relevance values capture pertinent information with respect to the classification problem. Indeed in all deletion experiments, we observe the most pregnant decrease resp.\ increase of the classification accuracy when using LRP as relevance model.
We additionally note that SA, in contrast to LRP,  is largely unable to provide suitable information linking to words that speak {\it against} a specific classification decision. Instead it appears that  the lowest SA relevances (which mainly correspond to zero-valued relevances) are more likely to identify words that have no impact on the classifier decision at all, as this deletion scheme has even less impact on the classification performance than random deletion when deleting words in increasing order of their relevance, as shown by the second deletion experiment.

When confronting the different CNN models, we observe that the CNN2 and CNN3 models, as opposed to CNN1, produce a steeper decrease of the classification performance when deleting the most relevant words from the initially correctly classified documents, both when considering LRP as well as SA as relevance model, as shown by the first deletion experiment.
This indicates that the networks with greater filter sizes are more sensitive to single word deletions, most presumably because during these deletions the meaning of the surrounding words becomes less obvious to the classifier. This also provides some weak evidence that, while CNN2 and CNN3 behave similarly (which suggests that a convolutional filter size of two is already enough for the considered classification problem), the learned filters in CNN2 and CNN3 do not only focus on isolated words but additionally consider bigrams or trigrams of words, as their results differ a lot from the CNN1 model in the first deletion experiment.

\subsubsection{Quantifying the Explanatory Power}\label{sec:knn}
In order to quantitatively evaluate and compare the ML models in combination with a relevance decomposition or {\it explanation} technique, we apply the evaluation method described in section \ref{sec4:extrinsic_validation}. That is, we compute the accuracy of an external classifier (here KNN) on the classification of document summary vectors (obtained with the ML model's predicted class). For these experiments we remove test documents which are empty or contain only one word after preprocessing (this amounts to remove 25 documents from the 20Newsgroups test set). The maximum KNN mean accuracy obtained when varying the number of neighbors $K$ (corresponding to our EPI metric of explanatory power) is reported for several models and explanation techniques in Table~\ref{tab:20news_knn_CNNs}.

\begin{table}[!h]
\centering\small
\caption{Results averaged over 10 random data splits. For each semantic extraction method, we report the explanatory power index (EPI) corresponding to the {\it maximum} mean KNN accuracy obtained when varying the number of neighbors $K$, the corresponding standard deviation over the multiple data splits, and the hyperparameter $K$ that led to the maximum accuracy.}\label{tab:20news_knn_CNNs}

\begin{tabular}{l|l|c|l}
\hline
\multicolumn{2}{c|}{Semantic Extraction} 	& Explanatory Power Index (EPI) 	&	KNN parameter	\\\hline\hline
word2vec/CNN1  		& LRP (ew)  		& {\bf 0.8045} ($\pm$ 0.0044)		& $K=10$		\\
			& SA (ew)   		& 0.7924 ($\pm$ 0.0052) 		& $K=9$ 		\\
			& LRP   		& 0.7792 ($\pm$ 0.0047)			& $K=8$			\\
			& SA   	  		& 0.7773 ($\pm$  0.0041)		& $K=6$			\\
\hline
word2vec/CNN2  		& LRP (ew)  		& {\bf 0.8076}  ($\pm$  0.0041)		& $K=10$		\\
			& SA (ew)   		& 0.7993 	($\pm$  0.0045)		& $K=9$	 		\\
			& LRP   		& 0.7847 ($\pm$ 0.0043) 		& $K=8$			\\
			& SA   	  		& 0.7767  	($\pm$  0.0053)		& $K=8$			\\
\hline
word2vec/CNN3 		& LRP (ew)  		& {\bf 0.8034} ($\pm$  0.0039)		& $K=13$		\\
			& SA (ew)  		& 0.7931  ($\pm$  0.0048)		& $K=10$		\\
			& LRP   		& 0.7793 	($\pm$  0.0037) 	& $K=7$			\\
			& SA   	  		& 0.7739  ($\pm$  0.0054)		& $K=6$			\\
\hline
word2vec        	&  	TFIDF		& {\bf 0.6816} ($\pm$  0.0044)		& $K=1$			\\
			&    	uniform		&  0.6208  	($\pm$  0.0052)		& $K=1$			\\
\hline\hline
BoW/SVM			& LRP 			& {\bf 0.7978}  ($\pm$  0.0048)		& $K=14$		\\
			& SA   			& 0.7837  	($\pm$ 0.0047)		& $K=17$ 		\\
\hline
 BoW  			&  TFIDF		& {\bf 0.7592} ($\pm$  0.0039)		& $K=1$ 		\\
			&  uniform		& 0.6669 ($\pm$  0.0061)		& $K=1$ 		\\
\hline
\end{tabular}
\end{table}

When pairwise comparing the best CNN based weighting schemes with the corresponding TFIDF baseline result from Table~\ref{tab:20news_knn_CNNs}, we find that all LRP element-wise weighted combinations of word2vec vectors are statistical significantly better than the TFIDF weighting of word embeddings at a significance level of 0.05 (using a corrected resampled t-test \cite{Nadeau_2003}).
Similarly, in the bag-of-words space, the LRP combination of one-hot word vectors is significantly better than the corresponding TFIDF document representation with a significance level of 0.05. Lastly, the best CNN2 explanatory power index is significantly higher than the best SVM based explanation at a significance level of 0.10.

\begin{figure}[!h]
\centering
\includegraphics[width=1\textwidth]{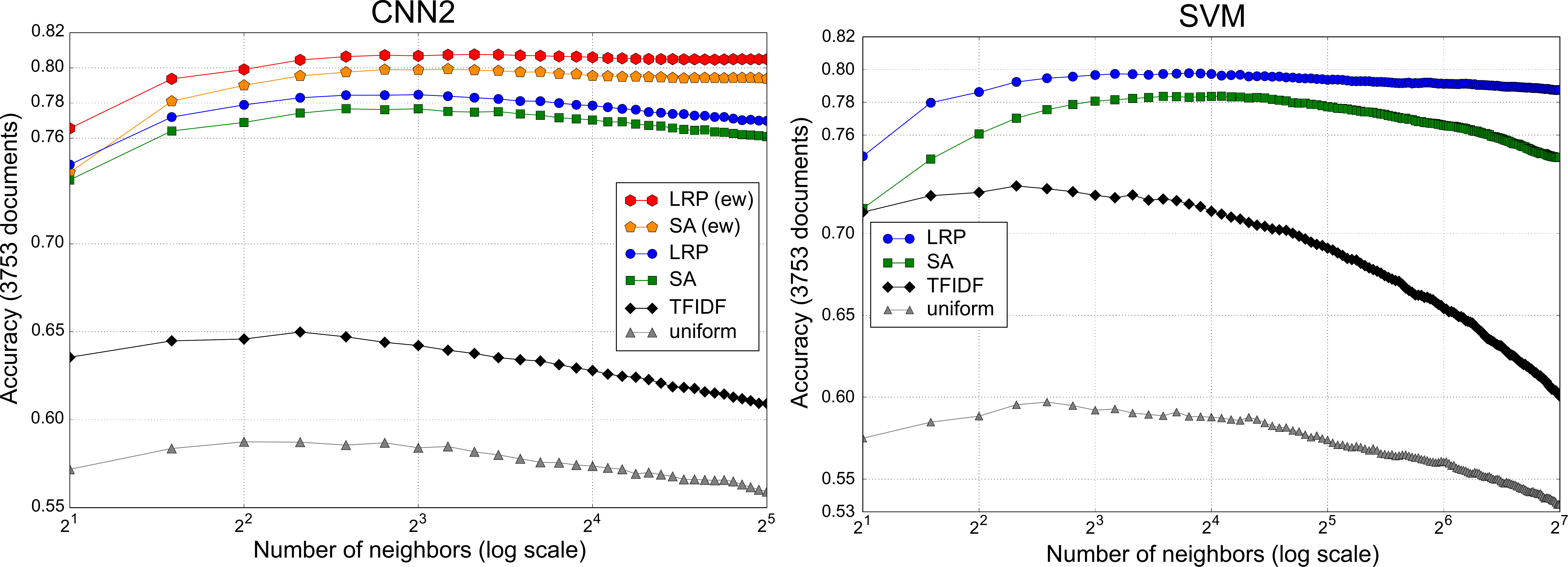}
\caption{KNN accuracy when classifying the document summary vectors of half of the 20Newsgroups test documents (other half is used as neighbors). Results are averaged over 10 random data splits.}\label{fig:20news_knn_CNN2}
\end{figure}

In Figure~\ref{fig:20news_knn_CNN2} we plot the mean accuracy of KNN (averaged over ten random test data splits) as a function of the number of neighbors $K$, for the CNN2 resp.\ the SVM model, as well as the corresponding TFIDF/uniform weighting baselines (for CNN1 and CNN3 we obtained a similar layout as for CNN2).
One can further see from Figure~\ref{fig:20news_knn_CNN2} that (1) (element-wise) LRP provides consistently better semantic extraction than all baseline methods and that (2) the CNN2 model has a higher explanatory power than the BoW/SVM classifier since it produces semantically more meaningful summary vectors for KNN classification.

Altogether the good performance, both qualitatively as well as quantitatively, of the element-wise combination of word2vec embeddings according to the LRP relevance illustrates the usefulness of LRP for extracting a new vector-based document representation presenting semantic neighborhood regularities in feature space, and let us presume other potential applications of relevance information, e.g.\ for aggregating word representations into sub-document representations like phrases, sentences or paragraphs.

\section{Conclusion}\label{sec6}

We have demonstrated qualitatively and quantitatively that LRP constitutes a useful tool, both for fine-grained analysis at the document level or as a dataset-wide introspection across documents, to identify words that are important to a classifier's decision. 
This knowledge enables to broaden the scope of applications of standard machine learning classifiers like support vector machines or neural networks, by extending the primary classification result with additional information linking the classifier's decision back to components of the input, in our case words in a document.
Furthermore, based on LRP relevance, we have introduced a new way of condensing the semantic information contained in word embeddings (such as word2vec) into a document vector representation that can be used for nearest neighbors classification, and that leads to better performance than standard TFIDF weighting of word embeddings.
The resulting document vector is the basis of a new measure of model explanatory power which was proposed in this work, and its semantic properties could beyond find applications in various visualization and search tasks, where the document similarity is expressed as a dot product between vectors. 

Our work is a first step toward applying the LRP decomposition to the NLP domain, and we expect this technique to be also suitable for various types of applications that are based on other neural network architectures such as character-based or recurrent network classifiers, or on other types of classification problems (e.g.\ sentiment analysis).
More generally, LRP could contribute to the design of more accurate and efficient classifiers, not only by inspecting and leveraging the input space relevances, but also through the analysis of intermediate relevance values at classifier ``hidden'' layers.

\section*{Acknowledgments}
This work was supported by the German Ministry for Education and Research as Berlin Big Data Center BBDC, funding mark 01IS14013A and by DFG. KRM thanks for partial funding by the National Research Foundation of Korea funded by the Ministry of Education, Science, and Technology in the BK21 program. \cl{Correspondence should be addressed to KRM and WS.}

\section*{Contributions}
Conceived the theoretical framework: LA, GM, KRM, WS. Conceived and designed the experiments: LA, FH, GM, KRM, WS. Performed the experiments: LA. Wrote the manuscript: LA, FH, GM, KRM, WS. Revised the manuscript:  LA, FH, GM, KRM, WS. Figure design:  LA, GM, WS. Final drafting: all equally.

{\small
\bibliography{bibliography}
}

\end{document}